\documentclass[sigconf]{acmart}

\usepackage{multirow}
\usepackage{pifont}

\makeatletter
\newcommand{\correspondingauthornote}{%
  \g@addto@macro\@authornotes{%
    \begingroup
      \renewcommand{\thefootnote}{\ding{41}}%
      \footnotetext{Corresponding author.}%
    \endgroup}}
\makeatother
\usepackage{makecell}
\usepackage{colortbl}

\definecolor{tableoursblue}{RGB}{234,244,250}
\definecolor{tablebandgray}{RGB}{242,243,245}

\AtBeginDocument{%
  }

\copyrightyear{2026}
\acmYear{2026}
\setcopyright{cc}
\setcctype{by}
\acmConference[MM '26]{Proceedings of the 34th ACM International Conference on Multimedia}{November 10--14, 2026}{Rio de Janeiro, Brazil}
\acmBooktitle{Proceedings of the 34th ACM International Conference on Multimedia (MM '26), November 10--14, 2026, Rio de Janeiro, Brazil}
\acmDOI{10.1145/3767308.3834996}
\acmISBN{979-8-4007-2213-4/2026/11}
\begin{document}

\title{Observe Less, Understand More: Cost-aware Cross-scale Observation for Remote Sensing Understanding}

\author{Zhenghao Xie}
\orcid{0009-0000-3634-4372}
\email{xiehaoai@whu.edu.cn}
\affiliation{%
  \department{School of Artificial Intelligence}
  \institution{Wuhan University}
  \city{Wuhan}
  \country{China}}

\author{Jing Xiao}
\correspondingauthor
\correspondingauthornote

\orcid{0000-0002-0833-5679}
\email{jing@whu.edu.cn}
\affiliation{%
  \department{School of Artificial Intelligence}
  \institution{Wuhan University}
  \city{Wuhan}
  \country{China}}

\author{Zhenqi Wang}
\orcid{0009-0003-2237-5997}
\author{Kexin Ma}
\orcid{0009-0000-8063-7316}
\affiliation{%
  \department{School of Artificial Intelligence}
  \institution{Wuhan University}
  \city{Wuhan}
  \country{China}}

\author{Liang Liao}
\orcid{0000-0002-2238-2420}
\affiliation{%
  \department{Hangzhou Institute of Technology}
  \institution{Xidian University}
  \city{Hangzhou}
  \country{China}}

\author{Gui-song Xia}
\orcid{0000-0001-7660-6090}
\affiliation{%
  \department{School of Artificial Intelligence}
  \institution{Wuhan University}
  \city{Wuhan}
  \country{China}}

\author{Mi Wang}
\orcid{0000-0003-2799-5987}
\affiliation{%
  \department{Hubei Luojia Laboratory}
  \institution{Wuhan University}
  \city{Wuhan}
  \country{China}}
\renewcommand{\shortauthors}{Zhenghao Xie et al.}

\begin{abstract}
Remote sensing understanding inherently requires multi-resolution observation, since different targets and application tasks demand different levels of spatial detail. While low-resolution (LR) imagery enables efficient global observation, high-resolution (HR) imagery provides critical local details at a much higher acquisition cost and with limited coverage. This motivates a cross-scale sensing strategy that selectively acquires HR imagery guided by LR-based global perception to improve task performance under constrained cost.
Existing HR sampling methods typically make selection decisions from isolated LR patches, thereby ignoring fine-grained intra-patch importance and cross-patch contextual interactions, leading to fragmented feature representation and suboptimal scene reasoning under sparse HR observations. To address this issue, we formulate cross-scale remote sensing understanding as a unified cost-aware problem that couples fine-grained HR sampling with cross-patch representation prediction, enabling more effective task reasoning with fewer HR observations.
Furthermore, we present GL-10M, a high- and low-resolution dataset with nearly 100,000 scene pairs and 10 million images for large-scale cross-resolution pretraining. Extensive experiments on recognition and retrieval tasks show that our method consistently achieves a superior performance--cost trade-off.
The code is publicly available at \url{https://github.com/xzhacc/CrossSO}.
\end{abstract}

\ccsdesc[300]{Computing methodologies~Image representations}
\ccsdesc[500]{Computing methodologies~Scene understanding}
\ccsdesc[500]{Information systems~Multimedia and multimodal retrieval}

\keywords{Remote Sensing, Cross-Scale Understanding, Sparse Observation}

\maketitle

\begin{figure*}[t]
  \centering
  \begin{minipage}[t]{0.31\linewidth}
    \centering
    \includegraphics[width=\linewidth]{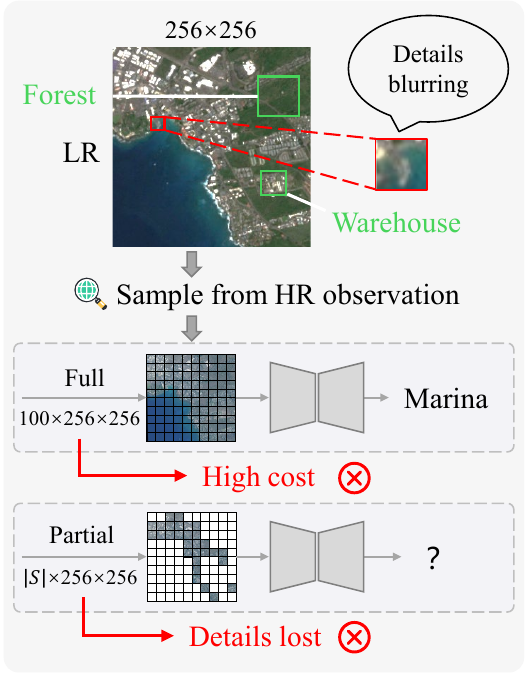}\\
    (a) Motivations
  \end{minipage}\hfill
  \begin{minipage}[t]{0.31\linewidth}
    \centering
    \includegraphics[width=\linewidth]{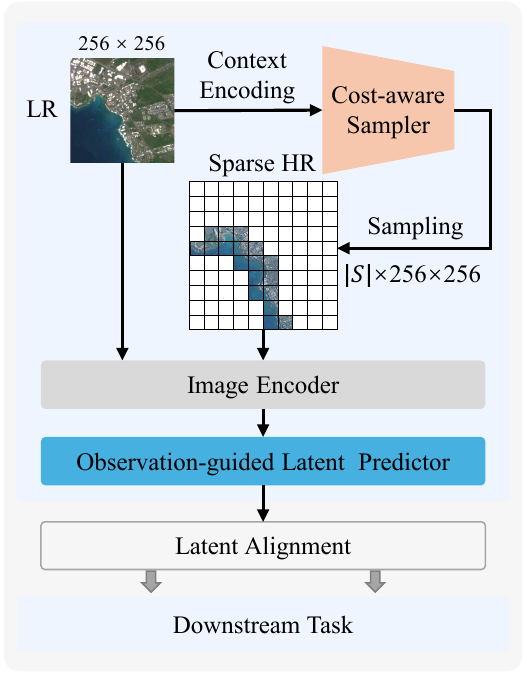}\\
    (b) Our method
  \end{minipage}\hfill
  \begin{minipage}[t]{0.31\linewidth}
    \centering
    \includegraphics[width=\linewidth]{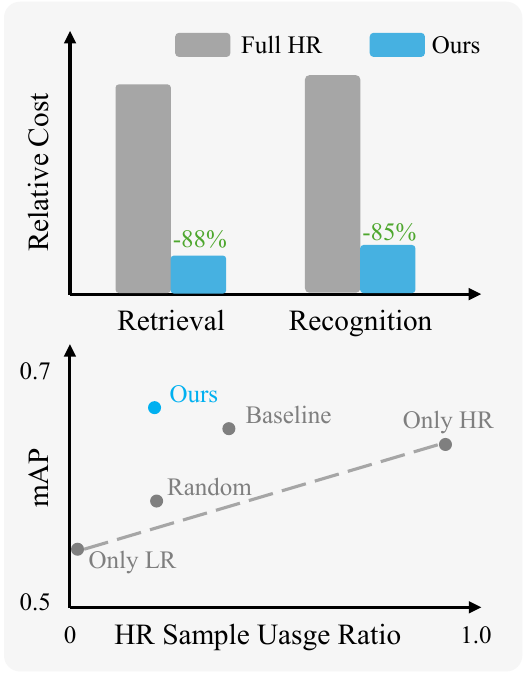}\\
    (c) Performance
  \end{minipage}
  \caption{\textbf{Cost-aware cross-scale understanding with sparse HR observations.} (a) Dense HR preserves details at high cost, whereas sparse HR leaves incomplete scene evidence. (b) From an LR overview, the cost-aware sampler selects HR tiles, and the observation-guided latent predictor completes the representation for recognition and retrieval. (c) Relative to full HR, our method reduces the observed HR area by 86\% on average across both tasks while maintaining comparable or better performance.}
  \Description{A three-panel overview contrasts dense and sparse high-resolution observation, illustrates the proposed low-resolution-guided selection and completion framework, and summarizes its performance and cost advantages.}
  \label{fig:intro}
\end{figure*}

\section{Introduction}
\label{sec:intro}

Remote sensing foundation models increasingly support diverse visual and multimodal Earth-observation tasks~\cite{huang2025survey}. Remote sensing applications fundamentally depend on multi-resolution observation, since different targets and application tasks require different levels of spatial detail. Large-area scene perception often relies on low-resolution (LR) imagery for efficient global coverage, whereas accurate recognition of small objects and subtle structures requires high-resolution (HR) imagery. As a result, effective remote sensing understanding can be achieved by integrating both global context and fine-grained local evidence across resolutions.

Recent Earth-observation foundation models have strengthened multi-sensor and multimodal representations~\cite{Guo_2024_SkySense,Zhang_2025_SkySenseV2,Nedungadi_2024_MMEarth,Astruc_2024_OmniSat}. AnySat further unifies representations across resolutions, scales, and modalities~\cite{Astruc_2025_AnySat}. Remote-sensing vision-language models support language-aligned recognition and retrieval~\cite{liu2024remoteclip,zhang2024rs5m,mall2024remote}. A recent survey covers this broader landscape~\cite{weng2025visionlanguage}, while instruction-based models extend it to interaction and visual grounding~\cite{Hu_2023_RSGPT,Kuckreja_2024_GeoChat,pang2025vhm}. Other assistants and ecologically supervised models broaden the available forms of semantic knowledge~\cite{Muhtar_2024_LHRSBot,Soni_2025_EarthDial,Daroya_2025_WildSAT}. However, these advances do not remove the acquisition cost of HR imagery. In practical systems, HR imagery may be acquired selectively rather than exhaustively, so multi-resolution observations differ not only in semantic content but also in cost and availability. HR imagery is therefore a scarce but informative signal that should be selectively acquired and efficiently utilized. This shifts the problem from passive multi-resolution fusion toward cost-aware cross-scale observation and understanding, where models have to reason about what to observe and at which resolution under constrained observation budgets.

Recent work has explored selective HR observation under resource constraints~\cite{uzkent2020learning,Revankar2024Scale,liu2025zoomearth}. These approaches focus on deciding which regions should be inspected at finer resolution under a limited budget, leaving the remaining regions without HR evidence. For recognition and retrieval, however, the representation must still account for the missing HR information. This motivates coupling selective observation with latent prediction to infer task-aligned representations at unobserved HR positions.

To address this gap, we propose a unified framework for cost-aware cross-scale remote sensing understanding, as illustrated in Figure~\ref{fig:intro}. The key idea is to integrate informative HR tile selection with observation-guided latent prediction in a unified inference framework. The predictor builds on the joint-embedding predictive architecture~\cite{assran2023ijepa}, inferring task-aligned representations at unobserved HR positions from the global LR context and sparse HR evidence rather than reconstructing pixels. Given an LR global observation, the sampler selects a small number of HR tiles using joint structural and semantic cues. The predictor then completes the unobserved positions in latent space. The resulting representation supports remote sensing recognition and retrieval while keeping HR acquisition tied to the available observation budget.

Our main contributions are summarized as follows:
\begin{itemize}
    \item We cast cross-scale remote sensing understanding as a\linebreak budget-constrained observation problem, explicitly modeling HR acquisition as a resource-aware selection problem under limited coverage.
    \item We propose a unified cost-aware framework that integrates informative HR tile selection with observation-guided latent completion, producing a task-aligned representation from sparse HR observations and global LR context.
    \item We construct GL-10M, a high- and low-resolution dataset with nearly 100,000 scene pairs and 10 million images for large-scale pretraining. Experiments across multiple recognition and retrieval benchmarks show that the framework achieves comparable or better performance than competing baselines while observing as little as 12.3\% of the HR area.
\end{itemize}

\section{Related Work}

\subsection{Cross-scale Representation Learning}
Cross-scale representation learning addresses the semantic mismatch between images acquired at different spatial resolutions. Some methods model concept scale or resource limits for recognition and reinforcement learning~\cite{Revankar2024Scale,Liu_2024_SAN}, while masked pretraining and domain adaptation learn across resolutions~\cite{Reed_2023_ScaleMAE,Liu_2024_MSAR}. Large-scale geospatial pretraining covers variations in sensors, seasons, and spatial resolution~\cite{Bastani_2023_Satlas,Manas_2021_SeasonalContrast,Sun_2022_RingMo,Fuller_2022_SatViT}. Temporal changes can guide region sampling~\cite{Mall_2023_ChangeAware}; cross-resolution consistency supports vehicle counting in very-low-resolution imagery~\cite{zhao2022vehiclecounting}; and historical images guide multi-temporal rescaling~\cite{zhang2024refscale}. Together, these studies show how complementary spatial or temporal evidence can strengthen coarse observations. Incomplete multimodal segmentation addresses missing sensor inputs~\cite{wen2026sgma}, whereas budgeted cross-scale observation introduces spatially structured missingness: the full scene is visible at LR, and HR evidence is available only in selected regions. Joint-embedding prediction estimates target representations from visible context without reconstructing pixels~\cite{assran2023ijepa}. We apply this idea to budgeted cross-scale observation, using global LR context and sparse HR tiles to estimate representations at unobserved HR regions.

\subsection{Efficient High-resolution Perception}
High-resolution perception incurs costs throughout acquisition, transmission, and downstream processing. Window-based or selective attention and token pruning reduce processing cost within image models~\cite{Liu_2021_Swin,li2024flexattention,Kong_2022_ECCV}. Vision-language model efficiency methods remove redundant visual tokens, reduce encoding overhead, or distill smaller networks~\cite{Chen_2024_FastV,Vasu_2025_FastVLM,Alvar_2025_DivPrune,Yang_2025_TopV,Jang_2025_VL2Lite}. At the system level, historical-reference compression reduces satellite downlink demand~\cite{wang2023uplink}, while resource-limited on-orbit platforms balance computing, storage, and real-time transmission~\cite{wang2024luojia}. These lines of work improve efficiency after imagery has been acquired. Active observation instead addresses the acquisition decision itself. Active Zooming selects HR patches from an LR view, while IS-Count samples geographic regions for aggregate counting~\cite{uzkent2020learning,Meng_2022_ISCount}. Scale-aware Recognition uses cross-resolution disagreement to allocate HR observations; ZoomEarth adapts cropping and zooming to ultra-high-resolution geospatial vision-language tasks~\cite{Revankar2024Scale,liu2025zoomearth}. Although their tasks differ, each method allocates HR budgets through selective observation. Our framework likewise selects HR evidence, then applies cross-scale latent completion to unobserved positions. The predictor combines the LR overview with the selected HR tiles to form a completed scene representation under a limited HR budget.

\section{Problem Formulation}
\label{sec:problem_formulation}

Dense HR imagery is expensive to acquire and process for large-area remote sensing understanding. We formulate the task as cross-scale observation under a budget constraint. Let $I^{g}$ be an LR overview, and divide the corresponding HR scene $I^{h}=\{I^{h}_{p}\}_{p=1}^{P}$ into $P$ equal-area candidate tiles. We use $P=100$, arranged as a $10\times10$ grid over the same geographic extent. Each HR tile has the same pixel dimensions as $I^{g}$. The full HR scene therefore has $10\times$ the linear resolution of the LR overview. Based on $I^{g}$, the observation policy acquires a subset $S\subseteq\{1,\ldots,P\}$. Because the tiles cover equal areas, $|S|/P$ is our normalized measure of HR acquisition cost. The average budget is denoted by $\rho$, and the realized dataset-level mean is reported as OBR.

After encoding, $F^{g}$ denotes the LR feature and $F^{h}_{p}$ the feature of HR tile $p$. We retain the tile locations in $F^{h}_{S}=\{(p,F^{h}_{p})\mid p\in S\}$. From $I^{g}$, the sampler predicts a normalized score map $\hat{\mathbf{G}}=\pi(I^{g})\in[0,1]^{P}$. Applying a shared threshold calibrated to $\rho$ gives $S=\operatorname{Select}_{\rho}(\hat{\mathbf{G}})$, with the number of selected tiles varying across scenes. The completion function combines the LR context and sparse HR evidence as $\tilde{H}=\phi(F^{g},F^{h}_{S})$. This representation covers the full spatial grid, retaining the observed HR evidence and filling the unobserved positions in latent space.

For a downstream target $y$, let $\mathcal L_{\mathrm{task}}(\tilde H,y)$ include either a learned task head or a fixed similarity-based inference rule. The system objective is
\begin{equation}
\min_{\pi,\phi}\;
\mathbb{E}_{(I^{g},I^{h},y)\sim\mathcal D}
\left[\mathcal L_{\mathrm{task}}(\tilde H,y)\right]
\quad
\mathrm{s.t.}\quad
\mathbb{E}_{I^{g}\sim\mathcal D}
\left[\frac{|S|}{P}\right]\leq\rho.
\end{equation}
Because the expectation is taken over $\mathcal D$, $\rho$ is a dataset-level operating point rather than a per-scene quota. The policy may therefore vary $|S|$ across scenes, while gains at a matched $\rho$ reflect how effectively the available HR evidence is selected and used to complete the scene representation. Since HR encoding is applied only to the selected tiles, the same constraint also limits the HR-dependent computation. It provides a common system-level criterion for the sampler and predictor, whose practical training objectives are specified next.

\section{Methodology}
\subsection{Framework Overview}

\begin{figure*}[t]
  \centering
  \includegraphics[width=0.98\linewidth]{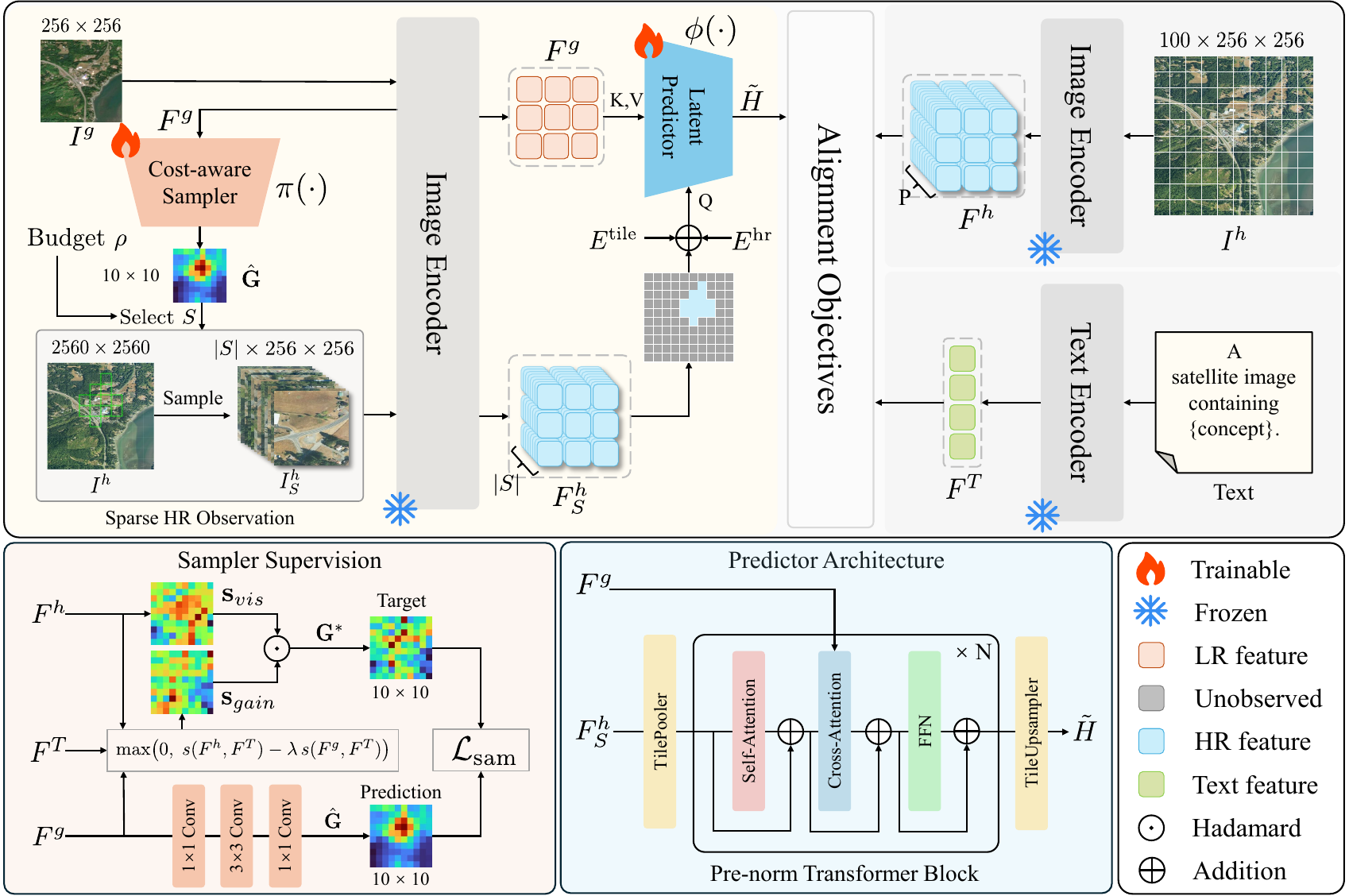}
  \caption{\textbf{Cost-aware sampling and latent completion for cross-scale understanding.} Given an LR overview $I^{g}$, the sampler selects informative HR tiles. The observation-guided latent predictor combines the selected HR evidence with global LR context to produce a completed representation $\tilde{H}$. Supervised tasks use dense visual targets. For zero-shot tasks, $\tilde{H}$ can be aligned directly with text features or with visual features already aligned to the same text space. The lower panels show the sampler and predictor in detail.}
  \label{fig:framework}
  \Description{The framework first selects high-resolution regions from a low-resolution global overview, then completes a scene representation from sparse local evidence and global context, and finally aligns it with task-specific target features for supervised and zero-shot tasks.}
\end{figure*}

Figure~\ref{fig:framework} summarizes the observation policy $\pi$ and completion function $\phi$ defined in Section~\ref{sec:problem_formulation}. Given an LR overview $I^{g}$, the cost-aware sampler scores the HR tiles and selects $S$ at the target operating point. The predictor combines $F^{h}_{S}$ with $F^{g}$ into $\tilde{H}$ for recognition and retrieval. The visual and textual alignment targets in the figure are instances of the abstract task target $H^{\star}$.

Training has two stages. In Stage I, the predictor learns to recover full-scene target representations from random budgeted observations, while the target encoder remains frozen. Random masks expose it to varied observation patterns without coupling its early training to a particular sampling policy. In Stage II, task-specific supervision is applied to $\tilde{H}$, and the sampler is trained separately from offline paired LR and HR supervision. Supervised recognition and retrieval use dense visual targets. Zero-shot tasks use a shared image-text space, where $\tilde{H}$ can be supervised either by text embeddings or by visual embeddings from an image encoder already aligned with text. During inference, the sampler receives only $I^{g}$. A shared threshold on its normalized scores controls the HR observation cost while allowing the number of selected tiles to vary across scenes.

\subsection{Cost-aware Cross-scale Sampler}
Given $I^{g}$, the sampler must distinguish tiles that are merely detailed from those whose HR observations add useful information. A visually complex region may already be well described by the LR overview. Conversely, a less conspicuous region may contain concept-level evidence that is missing at LR. We therefore use two complementary cues: structural saliency estimates where fine spatial content is difficult to recover, whereas cross-scale semantic gain asks whether an HR observation adds concept-level evidence beyond the LR view.

\subsubsection{Structural Saliency Prior} We derive the structural saliency score $s_{\mathrm{vis}}(p)$ from dense features produced by a frozen visual encoder, such as DINOv2~\cite{oquab2023dinov2}. Tiles receive higher scores when their features contain stronger local variation, sharper boundaries, or more complex spatial patterns. Such details are easily lost at LR and are difficult to infer from surrounding context when the corresponding HR tile is not observed.

\subsubsection{Cross-scale Semantic Gain} A structurally salient tile may still be semantically redundant with the LR overview. We measure the additional open-vocabulary evidence provided by HR tile $I^{h}_{p}$ using
\begin{equation}
s_{\mathrm{gain}}(p)=\max\!\left(0,\; s(F^{h}_{p},F^{T})-\lambda\, s(F^{g},F^{T})\right),
\end{equation}
where $F^{g}$, $F^{h}_{p}$, and $F^{T}$ are extracted in the same frozen vision-language space. The function $s(\cdot,\cdot)$ measures feature similarity, and $\lambda$ is the LR-semantic discount factor, fixed to $1.0$ in all experiments. The LR term is shared across the candidate tiles of a scene, so the score discounts concepts that are already evident from the overview. A high value therefore indicates that the HR tile resolves semantics that remain weak or ambiguous in $I^{g}$.

We take the Hadamard product of $s_{\mathrm{vis}}$ and $s_{\mathrm{gain}}$, followed by global normalization, to construct the supervision map $\mathbf{G}^{*}$. The product suppresses tiles that are strong under only one cue and favors regions that are both difficult to infer and semantically useful. Both cue maps are computed offline from paired LR and HR data. The sampler is not optimized through the downstream task loss because tile selection is discrete and a strong predictor further weakens the signal reaching the sampling decisions. Instead, it learns the offline map directly by predicting $\hat{\mathbf{G}}=\pi_{\theta_{\mathrm{sam}}}(I^{g})$ from the LR input and uses binary cross-entropy as the sampler loss, $\mathcal{L}_{\mathrm{sam}}=\operatorname{BCE}(\hat{\mathbf{G}},\mathbf{G}^{*})$. The frozen encoders used to build $\mathbf{G}^{*}$ are not required at deployment. A shared threshold calibrated to the target observation ratio $\rho$ converts the predicted scores into $S$. Because the threshold does not impose a fixed tile count per scene, the realized allocation can contract on simple scenes and expand on structurally complex ones while meeting the budget on average. The resulting observation set therefore reflects both inference difficulty and task relevance, and becomes the sparse local evidence supplied to the latent predictor. Because the score map remains aligned with the HR grid, each acquisition decision is spatially traceable, enabling tile-level inspection of the realized allocation.

\subsection{Observation-guided Latent Predictor}
Most HR tiles remain missing after sampling. Rather than reconstructing their pixels, the predictor estimates the missing scene features from the selected HR evidence and the LR overview. This follows the general idea of predicting target representations from visible context~\cite{assran2023ijepa}, but applies it to cross-scale observations. Working in feature space avoids spending capacity on exact low-level texture synthesis. The HR features provide reliable local detail, while the LR feature supplies scene-level context for the unobserved regions.

\subsubsection{Architecture of the Predictor} The observed features $F^{h}_{S}$ are placed back on the HR grid, and a shared learnable mask embedding fills the unobserved locations. Rasterization retains the acquisition pattern rather than reducing the evidence to an unordered set. Here, $\mathbf{M}(F^{h}_{S})$ is the rasterized sparse HR feature map, while $E^{\mathrm{tile}}$ and $E^{\mathrm{hr}}$ encode spatial positions at two levels. The predictor processes the entire structured sparse map with 12 stacked cross-attention blocks. Cross-attention uses HR queries and global LR keys and values. The overall mapping is
\begin{equation}
\tilde{H} = \phi_{\theta_{\mathrm{pred}}}\!\left(F^{g},\, \operatorname{TilePool}\!\left(\mathbf{M}(F^{h}_{S}) + E^{\mathrm{tile}} + E^{\mathrm{hr}}\right)\right).
\end{equation}
The selected HR tiles indicate what local evidence is available; the LR overview constrains how the missing features should fit the scene as a whole. An upsampling layer after the cross-attention stack restores the target feature resolution.

\subsubsection{Latent Bottleneck}
Heavy masking makes direct prediction on the full HR grid expensive and difficult to optimize. We apply $\operatorname{TilePool}$ before the cross-attention stack to reduce the number of HR tokens. The predictor completes the representation in this compact latent space, after which the tokens are upsampled to the target resolution. Besides reducing computation, the bottleneck encourages scene-level completion instead of copying or memorizing isolated sparse patterns.

\subsubsection{Dual Positional Encoding} The predictor needs to retain both the acquisition location of each tile and the finer layout within the HR scene. The tile-level embedding $E^{\mathrm{tile}}$ identifies the coarse region to which a feature belongs, anchoring each observation to the sampling grid. The HR-level embedding $E^{\mathrm{hr}}$ records its finer spatial position at the target resolution. Using both embeddings preserves the geometry of the sparse observations through pooling and subsequent upsampling.

\subsubsection{Predictor Training} A direct regression objective can admit low-variance predictions that fit the average target while discarding scene structure. We therefore use a frozen target encoder $\Psi$ to extract the full-observation representation $H^{\star}$ and stop gradients through this branch. LayerNorm is applied separately to the predicted and target features before the loss is computed. To keep the notation compact, the normalized features retain the same symbols below. We subtract the same target-feature center from both representations and optimize
\begin{equation}
\mathcal{L}_{\mathrm{pred}}
=
1
-
\frac{1}{|\mathcal{T}|}
\sum_{j \in \mathcal{T}}
\frac{
\left\langle \tilde{H}_{j}-\mu,\; H^{\star}_{j}-\mu \right\rangle
}{
\|\tilde{H}_{j}-\mu\|_{2}\,\|H^{\star}_{j}-\mu\|_{2}
},
\end{equation}
where $\mathcal{T}$ is the set of target tokens. The center $\mu$ is initialized from the target features of the first distributed training batch and then kept fixed. Apart from a constant scale, the cosine-regression term has the same normalized latent-regression form used in BYOL~\cite{grill2020byol}; the fixed centering is specific to our implementation. Stage I also applies an auxiliary cosine-regression loss $\mathcal{L}_{\mathrm{ctx}}$ to the context tokens and optimizes it jointly with $\mathcal{L}_{\mathrm{pred}}$.

\subsection{Latent Alignment Objectives}
The target space depends on the downstream task, but the sampler and predictor interface remains unchanged. Supervised recognition and retrieval retain the token-level structure of dense visual features. Zero-shot tasks instead use a pooled representation in a shared image-text space. In the latter case, supervision can come from text itself or from visual features that have already been aligned with text.

\subsubsection{Dense Visual Latent Alignment}
For supervised recognition and retrieval, $\Psi$ is a frozen dense visual encoder that extracts $H^{\star}$ from the complete HR observation. The predictor uses $\mathcal{L}_{\mathrm{pred}}$ to recover this token-level target from the LR overview and sparse HR observations. Unlike a globally pooled target, $H^{\star}$ preserves the ordered HR token layout, so completion is constrained at every spatial position. Because the loss is applied in feature space, the predictor retains the spatial organization and cross-scale consistency of the target without reconstructing image pixels. The predictor is then fine-tuned with downstream labels.

\subsubsection{Vision-language Latent Alignment}
For zero-shot recognition and retrieval, we pool $\tilde{H}$ and place it in a shared image-text space~\cite{Radford_2021_CLIP}. This branch accepts two types of targets. The direct route uses text embeddings produced by a frozen text encoder. The visual-intermediary route instead uses features from a frozen image encoder whose representation has already been aligned with the same text space. The latter follows the idea used in GRAFT~\cite{mall2024remote}: language capability can be transferred through an aligned visual space rather than requiring the predictor to regress class names directly. In either route, training pulls matched scene and target representations together and separates non-matching targets within the minibatch. The target encoders remain frozen, so both routes optimize only the completion pathway against a stable semantic space. At inference, class-name embeddings from the paired text encoder are matched against the completed representation; thus the text encoder remains the query interface even when aligned visual features supply the training target.

\begin{table*}[t]
  \centering
  \caption{Main comparison of retrieval-based recognition for concepts \textbf{seen} and \textbf{unseen} during training on Sentinel-2/NAIP. Results include retrieval performance, HR observation usage, and inference efficiency. N.L. denotes nightlight sampling; bold and underlined values mark the best and second-best results per setting, respectively; light blue rows mark our methods.}
  \label{tab:retrieval_zeroshot}
  \begin{minipage}{\textwidth}
  \centering
  \small
  \setlength{\tabcolsep}{3.6pt}
  \renewcommand{\arraystretch}{1.08}
  \begin{tabular}{
    >{\raggedright\arraybackslash}p{0.288\textwidth}
    >{\centering\arraybackslash}p{0.066\textwidth}
    *{6}{>{\centering\arraybackslash}p{0.0886\textwidth}}
    }
    \toprule
    \multirow{2}{*}[-0.55ex]{Method} & \multirow{2}{*}[-0.55ex]{Input} &
    \multicolumn{2}{c}{Seen Concepts} & \multicolumn{2}{c}{Unseen Concepts} &
    \multirow{2}{*}{\makecell{HR Usage\\(OBR)$\downarrow$}} &
    \multirow{2}{*}{\makecell{Time\\(s)$\downarrow$}} \\
    \cmidrule(lr){3-4} \cmidrule(lr){5-6}
    & & mAP@100$\uparrow$ & mAP@20$\uparrow$ & mAP@100$\uparrow$ & mAP@20$\uparrow$ & & \\
    \midrule
    \rowcolor{tablebandgray}\textit{Full observation} & & & & & & & \\
    GRAFT (HR)~\cite{mall2024remote,Revankar2024Scale} & HR & 0.501 & 0.513 & \underline{0.541} & \underline{0.574} & 1.000 & \textbf{962} \\
    GRAFT (LR+LLM)~\cite{mall2024remote,Revankar2024Scale} & LR+HR & \underline{0.530} & \underline{0.561} & \textbf{0.549} & 0.573 & 1.000 & 984 \\
    \rowcolor{tableoursblue}Ours-Full & LR+HR & \textbf{0.589} & \textbf{0.602} & 0.539 & \textbf{0.592} & 1.000 & \underline{968} \\
    \specialrule{0.4pt}{1pt}{0pt}
    \rowcolor{tablebandgray}\textit{LR only} & & & & & & & \\
    CLIP-RSICD~\cite{Arutiunian_2021_CLIPRSICD} & LR & 0.393 & 0.441 & 0.311 & 0.303 & 0.000 & 11 \\
    OpenCLIP~\cite{Cherti_2023_ScalingCLIP} & LR & 0.481 & 0.505 & 0.392 & 0.447 & 0.000 & \textbf{10} \\
    GRAFT (KD)~\cite{mall2024remote,Revankar2024Scale} & LR & \textbf{0.534} & \textbf{0.559} & \textbf{0.490} & \textbf{0.503} & 0.000 & 11 \\
    RemoteCLIP~\cite{liu2024remoteclip} & LR & 0.461 & 0.493 & 0.413 & 0.440 & 0.000 & 11 \\
    \rowcolor{tableoursblue}Ours-LR & LR & \underline{0.498} & \underline{0.531} & \underline{0.468} & \underline{0.495} & 0.000 & \textbf{10} \\
    \specialrule{0.4pt}{1pt}{0pt}
    \rowcolor{tablebandgray}\textit{Sparse observation} & & & & & & & \\
    GRAFT (LR+N.L. samp.)~\cite{mall2024remote,Meng_2022_ISCount,Revankar2024Scale} & LR+HR & 0.379 & 0.354 & 0.440 & 0.492 & \underline{0.236} & 323 \\
    Scale-aware (GRAFT)~\cite{Revankar2024Scale} & LR+HR & \underline{0.633} & \underline{0.639} & \underline{0.502} & \underline{0.564} & \underline{0.236} & \underline{233} \\
    \rowcolor{tableoursblue}Ours & LR+HR & \textbf{0.636} & \textbf{0.695} & \textbf{0.522} & \textbf{0.586} & \textbf{0.123} & \textbf{118} \\
    \bottomrule
  \end{tabular}
  \end{minipage}
\end{table*}

\section{Experimental Results}
\label{sec:exp}

\subsection{Benchmark Datasets}
\subsubsection{Sentinel-2/NAIP} This benchmark~\cite{Revankar2024Scale} pairs 10\,m Sentinel-2 imagery with 1\,m NAIP observations over the same geographic footprint. The LR image supplies a global view, while the NAIP observations provide local HR details. Forty OSM concepts span land cover, terrain, and built environments~\cite{Revankar2024Scale}; Geofabrik supplies the extracts~\cite{osm2024}. Following the original protocol, the concepts are divided into seen and unseen groups for retrieval-based recognition.

\subsubsection{GL-10M} We also introduce GL-10M, which extends the Sentinel-2/NAIP pairing to a substantially larger scale through spatial alignment and temporal matching. GL-10M contains nearly 100,000 scene pairs covering 66 geographic concepts in 11 U.S. states. Each pair includes one global LR image and 100 corresponding HR local images, giving approximately 10 million images in total. We use GL-10M for large-scale pretraining of the observation-guided latent predictor.
Dataset construction includes coverage and image-quality filtering, followed by geographically separated training, validation, and test splits.

\subsubsection{Zero-shot Benchmarks} Besides retrieval-based recognition on Sentinel-2/NAIP, we directly evaluate the GL-10M-pretrained model on EuroSAT~\cite{helber2019eurosat}, BigEarthNet~\cite{sumbul2019bigearthnet}, and a held-out GL-10M test split without further fine-tuning. EuroSAT and BigEarthNet test cross-dataset zero-shot transfer, whereas the GL-10M test split measures generalization to scenes excluded from pretraining. Recognition is performed by matching pooled visual features with text embeddings derived from class names.

\begin{table}[t]
  \centering
  \caption{Fully supervised retrieval-based recognition on Sentinel-2/NAIP under controlled HR observation settings.}
  \label{tab:retrieval_supervised}
  \small
  \setlength{\tabcolsep}{2.6pt}
  \renewcommand{\arraystretch}{1.08}
  \begin{tabular}{
    >{\raggedright\arraybackslash}p{0.39\columnwidth}
    c c c c
    }
    \toprule
    Method & Input & mAP@100$\uparrow$ & mAP@20$\uparrow$ & OBR$\downarrow$ \\
    \midrule
    \rowcolor{tablebandgray}\textit{Full observation} & & & & \\
    Scale-aware (HR)~\cite{Revankar2024Scale} & HR & 0.695 & 0.735 & 1.000 \\
    Scale-aware (LR+LLM)~\cite{Revankar2024Scale} & LR+HR & 0.689 & 0.731 & 1.000 \\
    Ours w/o LR & HR & \underline{0.718} & \textbf{0.769} & 1.000 \\
    \rowcolor{tableoursblue}Ours-Full & LR+HR & \textbf{0.722} & \underline{0.767} & 1.000 \\
    \specialrule{0.4pt}{1pt}{0pt}
    \rowcolor{tablebandgray}\textit{LR only} & & & & \\
    ResNet-50~\cite{He_2016_ResNet} & LR & 0.421 & 0.447 & 0.000 \\
    SatMAE~\cite{Cong_2022_SatMAE} & LR & 0.351 & 0.388 & 0.000 \\
    CSMAE~\cite{Tang_2023_CrossScaleMAE} & LR & 0.412 & 0.422 & 0.000 \\
    Scale-aware (LR)~\cite{Revankar2024Scale} & LR & 0.451 & 0.473 & 0.000 \\
    Scale-aware (KD)~\cite{Revankar2024Scale} & LR & \textbf{0.570} & \textbf{0.606} & 0.000 \\
    \rowcolor{tableoursblue}Ours-LR & LR & \underline{0.497} & \underline{0.554} & 0.000 \\
    \specialrule{0.4pt}{1pt}{0pt}
    \rowcolor{tablebandgray}\textit{Sparse observation} & & & & \\
    PatchDrop~\cite{uzkent2020learning,Revankar2024Scale} & LR+HR & 0.445 & 0.470 & 0.269 \\
    N.L. samp.~\cite{Meng_2022_ISCount,Revankar2024Scale} & LR+HR & 0.463 & 0.480 & \underline{0.236} \\
    Scale-aware (full)~\cite{Revankar2024Scale} & LR+HR & \underline{0.736} & \underline{0.796} & \underline{0.236} \\
    \rowcolor{tableoursblue}Ours & LR+HR & \textbf{0.755} & \textbf{0.811} & \textbf{0.146} \\
    \bottomrule
  \end{tabular}
\end{table}

\begin{table}[H]
  \centering
  \caption{Zero-shot recognition performance on EuroSAT~\cite{helber2019eurosat}, BigEarthNet~\cite{sumbul2019bigearthnet}, and GL-10M, where $\ast$ denotes support-weighted mAP for the long-tailed class distribution.}
  \label{tab:zeroshot_recognition}
  \small
  \setlength{\tabcolsep}{3.2pt}
  \renewcommand{\arraystretch}{1.08}
  \begin{tabular}{
    >{\raggedright\arraybackslash}p{0.28\columnwidth}
    >{\centering\arraybackslash}p{0.20\columnwidth}
    >{\centering\arraybackslash}p{0.235\columnwidth}
    >{\centering\arraybackslash}p{0.175\columnwidth}
    }
    \toprule
    \multirow{2}{*}[-0.25ex]{Method} & EuroSAT & BigEarthNet & GL-10M \\
           & Acc.$\uparrow$ & mAP$\uparrow$ & mAP$^{\ast}\uparrow$ \\
    \midrule
    CLIP~\cite{Radford_2021_CLIP} & 0.536 & 0.231 & 0.401 \\
    GRAFT~\cite{mall2024remote} & \underline{0.638} & \underline{0.325} & \underline{0.441} \\
    RemoteCLIP~\cite{liu2024remoteclip} & 0.386 & 0.228 & 0.362 \\
    \rowcolor{tableoursblue}Ours & \textbf{0.652} & \textbf{0.346} & \textbf{0.487} \\
    \bottomrule
  \end{tabular}
\end{table}

\subsection{Experimental Setup}
\subsubsection{Baselines}
\label{sec:baselines}
Comparisons are grouped by HR availability. Full observation covers the HR and LR+LLM variants of GRAFT~\cite{mall2024remote} and Scale-aware~\cite{Revankar2024Scale}, plus Ours w/o LR and Ours-Full; LR-only covers standard, remote-sensing, vision-language, and distilled encoders, including Scale-aware (LR), Scale-aware (KD), and Ours-LR; sparse observation covers PatchDrop~\cite{uzkent2020learning}, nightlight-sampled GRAFT~\cite{mall2024remote,Meng_2022_ISCount}, Scale-aware, and Ours. The first two groups define the availability endpoints, while sparse methods are assessed by both retrieval performance and OBR. Table~\ref{tab:zeroshot_recognition} evaluates cross-dataset zero-shot recognition with CLIP~\cite{Radford_2021_CLIP}, GRAFT, RemoteCLIP~\cite{liu2024remoteclip}, and Ours. Published baselines keep their original configurations and backbones; Ours variants are internal controls.

\subsubsection{Experimental Settings}
Unless otherwise specified, our models use ViT-B/16 as the visual backbone~\cite{Dosovitskiy_2020_ViT}. We pretrain the observation-guided latent predictor on GL-10M and fine-tune it only for the Sentinel-2/NAIP experiments in Tables~\ref{tab:retrieval_zeroshot} and~\ref{tab:retrieval_supervised}. Table~\ref{tab:zeroshot_recognition} directly evaluates the pretrained model. Training uses AdamW with an initial learning rate of $3\times10^{-4}$ and a weight decay of $10^{-2}$. A 32-epoch cosine-annealing schedule lowers the learning rate to $1\times10^{-5}$. Tables~\ref{tab:retrieval_zeroshot} and~\ref{tab:zeroshot_recognition} use class-name text embeddings at inference. Table~\ref{tab:retrieval_zeroshot} evaluates seen and unseen concepts on Sentinel-2/NAIP. Table~\ref{tab:retrieval_supervised} uses dense visual feature alignment for fully supervised retrieval-based recognition. Its comparison therefore changes the supervision target while keeping our backbone fixed. The baselines in Table~\ref{tab:zeroshot_recognition} retain their original backbones, so those results compare complete pretrained systems rather than controlled backbone variants. At inference, the sampler applies a shared threshold of 0.55 to globally normalized scores over the fixed $10\times10$ HR grid. At the reported operating point, the same threshold is applied to every test scene, allowing content-dependent tile counts under a common rule. We therefore report realized OBR when it differs from the nominal target.

\subsubsection{Evaluation Metrics}
For retrieval, we truncate each ranking at $K$ and average AP across queries to obtain mAP@$K$, where $K\in\{20,100\}$. We report top-1 accuracy on EuroSAT and mAP on BigEarthNet. GL-10M uses support-weighted mAP because of its long-tailed class distribution. Full-pipeline runtime includes sparse HR selection and downstream prediction. We measure it over the test set with a batch size of 32 on one NVIDIA RTX 4090 GPU. To measure observation cost, we define the \emph{Multi-scale Observation Cost Ratio (MOCR)}. Let $A$ denote the target area and $r_{\max}$ the highest available resolution. For requested observations $\mathcal{O}=\{(a_i,r_i)\}_{i=1}^{|\mathcal{O}|}$, with per-area cost $c(r)$, the requested and full-HR costs are $C_{\mathrm{req}}=\sum_i a_i c(r_i)$ and $C_{\mathrm{full}}=A c(r_{\max})$. Their ratio defines MOCR:
\begin{equation}
  \mathrm{MOCR} = \frac{C_{\mathrm{req}}}{C_{\mathrm{full}}}.
\end{equation}
Values below one indicate a cost reduction; values above one indicate redundant acquisition overhead. Because LR imagery is treated as nearly cost-free, HR acquisition dominates the cost in our experiments. MOCR therefore reduces to the \emph{Observation Budget Ratio (OBR)}. Under the equal-area HR partition, $\mathrm{OBR}(S)=A_{\mathrm{HR}}^{\mathrm{req}}/A=|S|/P$. The shared threshold allows the number of selected tiles to vary by sample, so OBR records the realized HR area. The parameter $\rho$ specifies the target operating point, not a fixed number of tiles per sample.

\subsection{Results}

\subsubsection{Main Results under Budgeted HR Observation}
\begin{figure}[t]
  \centering
  \includegraphics[width=\linewidth]{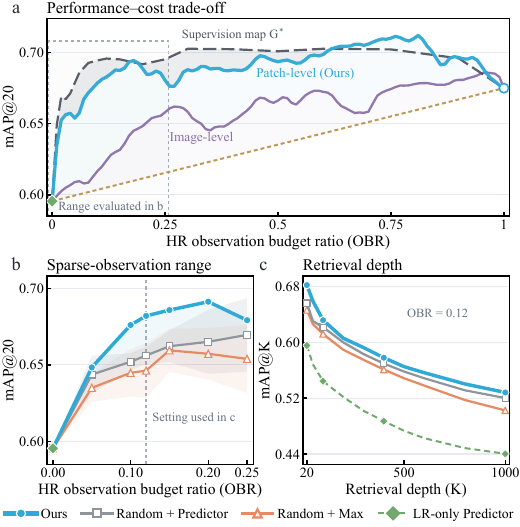}
  \caption{\textbf{Performance--cost trade-off, ablations, and retrieval robustness.} (a) $\mathbf{G}^{*}$ sampling provides the offline-supervision reference; patch-level supervision remains closer to it than image-level supervision. (b) Budget ablations isolate module effects. (c) Gains persist across depths at $\mathrm{OBR}=0.12$.}
  \label{fig:router_fixed}
  \Description{Three plots compare the cost--performance frontier, matched high-resolution observation budgets, and retrieval performance across depths at a fixed observation budget.}
\end{figure}

Following the scale-aware recognition protocol~\cite{Revankar2024Scale}, we formulate recognition as retrieval over geographic concepts and evaluate it under a limited HR observation budget. Figure~\ref{fig:router_fixed}(a) reports the results on the seen-concept split. Throughout the sparse-observation range, patch-level supervision produces a stronger performance-cost curve than image-level supervision and remains close to the $\mathbf{G}^{*}$ sampling reference. Their separation is largest at low OBR, which is also the regime most relevant to sparse acquisition. Table~\ref{tab:retrieval_zeroshot} reports the results at the selected operating point. Compared with Scale-aware (GRAFT), the strongest sparse-observation baseline, our method achieves relative gains of 0.5\% and 8.8\% in mAP@100 and mAP@20 on seen concepts, and 4.0\% and 3.9\% on unseen concepts. It also reduces OBR by 47.9\% and full-pipeline inference time by 49.4\%. On seen concepts, the larger gain in mAP@20 than in mAP@100 is consistent with a more pronounced improvement near the top of the retrieval ranking. The similar gains at both retrieval depths on unseen concepts show that the advantage is not tied to a particular value of $K$. With only 12.3\% of the HR area observed, the method outperforms all LR-only baselines and remains competitive with full-observation methods in the intended low-budget regime.

\subsubsection{Validation under Fully Supervised Setting}
Table~\ref{tab:retrieval_supervised} examines whether the same performance-cost pattern holds under fully supervised retrieval-based recognition. Compared with Scale-aware (full), the strongest sparse baseline in this setting, our method achieves relative gains of 2.6\% and 1.9\% in mAP@100 and mAP@20, respectively, while reducing OBR by 38.1\%. It also achieves higher performance than the LR-only and full-observation variants despite using substantially less HR input. The improvement is smaller than the top-ranking gain observed in the zero-shot setting, but it is obtained together with a clear reduction in observation cost. The performance-cost advantage is therefore retained when the supervision setting changes and is not confined to class-name-prompted zero-shot retrieval.

\subsubsection{Cross-dataset Zero-shot Generalization}
Table~\ref{tab:zeroshot_recognition} directly evaluates the model pretrained on GL-10M, without further fine-tuning, on EuroSAT~\cite{helber2019eurosat}, BigEarthNet~\cite{sumbul2019bigearthnet}, and the held-out GL-10M test split using class-name text prompts. Compared with GRAFT, the strongest baseline, our method obtains relative gains of 2.2\% on EuroSAT, 6.5\% on BigEarthNet, and 10.4\% on GL-10M. The gains are larger on the multi-label BigEarthNet and long-tailed GL-10M benchmarks than on EuroSAT. This pattern is consistent with a stronger benefit under more complex label structures and dataset distributions, although the cross-dataset comparison alone cannot isolate the cause of the difference. Together, these results show that the learned cross-scale representation generalizes to external datasets and held-out GL-10M scenes without target-specific adaptation.

\begin{table}[t]
  \centering
  \caption{Ablation of HR sampling and fusion on the unseen-concept split, reporting mAP@100 and OBR for each setting.}
  \label{tab:ablation}
  \small
  \setlength{\tabcolsep}{2.2pt}
  \renewcommand{\arraystretch}{1.08}
  \begin{tabular}{
    >{\raggedright\arraybackslash}p{0.29\columnwidth}
    >{\centering\arraybackslash}p{0.24\columnwidth}
    >{\centering\arraybackslash}p{0.22\columnwidth}
    >{\centering\arraybackslash}p{0.17\columnwidth}
    }
    \toprule
    \multicolumn{2}{c}{Settings} & \multicolumn{2}{c}{Metrics} \\
    \cmidrule(lr){1-2}\cmidrule(lr){3-4}
    HR Sampling & Fusion & mAP@100$\uparrow$ & OBR$\downarrow$ \\
    \midrule
    LR only & None & 0.468 & 0.000 \\
    \addlinespace[1pt]
    Random & Predictor & 0.495 & 0.123 \\
    \addlinespace[1pt]
    $s_{\mathrm{gain}}$ only & Predictor & 0.480 & 0.123 \\
    $s_{\mathrm{vis}}$ only & Predictor & 0.502 & 0.123 \\
    \addlinespace[1pt]
    $s_{\mathrm{vis}} \odot s_{\mathrm{gain}}$ & Max~\cite{Revankar2024Scale} & 0.475 & 0.123 \\
    \rowcolor{tableoursblue}
    $s_{\mathrm{vis}} \odot s_{\mathrm{gain}}$ & Predictor & \textbf{0.522} & 0.123 \\
    \bottomrule
  \end{tabular}
\end{table}

\begin{figure*}[t]
  \centering
  \includegraphics[width=0.97\linewidth]{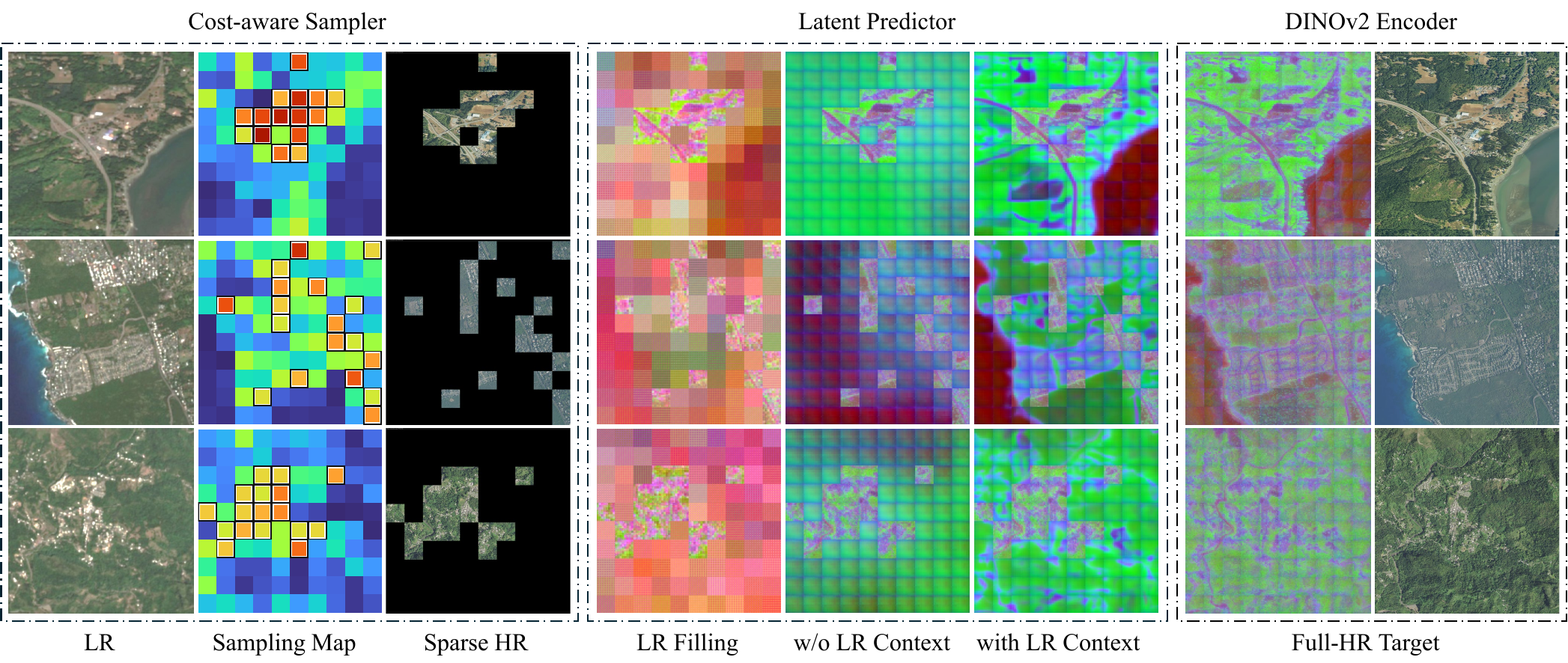}

    \caption{Qualitative comparison of cost-aware sampling and latent completion under sparse HR observation. The second column shows predicted sampling maps, while the latent feature visualizations share the same PCA basis and color scale. Using LR context produces more coherent scene-level predictions closer to the full-HR target.}
  \label{fig:predictor_vis}
  \Description{Three rows compare low-resolution inputs, predicted sampling maps, sparse high-resolution observations, latent completion variants, and full-resolution targets. Low-resolution context produces more coherent predictions.}
\end{figure*}

\subsection{Ablation Study}
\label{sec:ablation}
We evaluate the two components on the unseen-concept split in zero-shot retrieval under matched budgets to isolate their contributions. We then examine operating-point robustness across budgets and retrieval depths on the seen-concept split.

\subsubsection{Impact of the Cost-aware Cross-scale Sampler.}
As shown in Table~\ref{tab:ablation}, we fix fusion to the Predictor and vary only the HR sampling strategy. All sparse variants have an average OBR of 0.123. Across three independent trials, random sampling obtains $0.495 \pm 0.007$ mAP@100 (mean $\pm$ standard deviation). Joint sampling reaches 0.522 mAP@100, corresponding to relative gains of 5.5\% over random sampling and 4.0\% over structural saliency, the stronger single-score variant. Although semantic gain is less effective on its own, combining it with structural saliency yields the strongest retrieval result.
Figure~\ref{fig:predictor_vis} provides a direct view of the sampling behavior. Across the three scenes, the predicted maps assign higher scores to spatially heterogeneous or semantically informative regions. The sampler therefore concentrates its HR budget on selected regions instead of distributing it uniformly.

\subsubsection{Impact of the Observation-guided Latent Predictor.}
As shown in Table~\ref{tab:ablation}, we fix joint sampling and vary only the fusion method. At the same average OBR of 0.123, the Predictor achieves 0.522 mAP@100, a relative gain of 9.9\% over Max. Since the sampling strategy and observation cost remain unchanged, this comparison isolates the contribution of predictor-based latent completion.
Figure~\ref{fig:predictor_vis} compares LR filling, prediction without LR context, and the Predictor. In these examples, the Predictor yields feature maps whose spatial organization more closely matches the full-HR reference. Together with Table~\ref{tab:ablation}, these examples link LR context to more coherent spatial completion.

\subsubsection{Performance across Observation Budgets and Retrieval Depths.}
Figure~\ref{fig:router_fixed}(b) reports the seen-concept results across observation budgets. At each matched OBR, our method outperforms LR-only Predictor, Random+Max, and Random+Predictor in mAP@20. Since the methods are compared at the same OBR, the difference is not due to observing a larger area.
Figure~\ref{fig:router_fixed}(c) fixes the nominal target OBR at 0.12 (realized mean 0.123) and varies $K$. Our method remains ahead of the baselines across retrieval depths. Its advantage is therefore not tied to a single observation budget or a single value of $K$. Taken together, these analyses separate the roles of the two components: the sampler determines where the HR budget is spent, whereas the predictor uses the selected evidence and LR context to complete unobserved representations.

\section{Conclusion}
We presented a unified framework for cost-aware cross-scale remote sensing understanding that selects informative HR observations and uses sparse local evidence and global LR context to complete unobserved representations. We also constructed GL-10M, with nearly 100,000 HR/LR scene pairs and 10 million images for large-scale pretraining. Across multiple retrieval and recognition benchmarks, our method matched or outperformed full-observation, LR-only, and sparse-observation baselines while reducing HR observation cost by 86\% on average. These results support selective acquisition as a practical alternative to uniform HR coverage. Future work will study acquisition under model uncertainty and transfer across locations and time periods.

\begin{acks}
This work was supported by the National Natural Science Foundation of China (62371351), the Natural Science Foundation of Hangzhou (2025SZRJJ2224), the Key R\&D Program of Hubei Province (2025BEB011), and the Intelligent Computing Center of the National Cybersecurity Talent and Innovation Base, Wuhan.
\end{acks}

\bibliographystyle{ACM-Reference-Format}
\balance
\bibliography{references}


\clearpage
\nobalance
\raggedbottom
\section*{Supplementary Material}
This supplement describes the construction of GL-10M, examines LR-guided latent completion and threshold-controlled observation qualitatively, and reports implementation details and representative failure cases.

\subsection*{A. GL-10M Benchmark Details}

\subsubsection*{A.1 Data Sources and Pair Structure}
GL-10M combines Sentinel-2 imagery for the low-resolution (LR) branch, NAIP aerial imagery for the high-resolution (HR) branch, and OpenStreetMap (OSM) polygons for semantic supervision. Sentinel-2 and NAIP provide observations at 10\,m and 1\,m spatial resolution, respectively. Each scene contains one global LR image and 100 ordered local HR images over the same geographic extent. The HR observations form a fixed $10\times10$ grid, giving every tile an explicit spatial correspondence with the LR overview. We map OSM tags to 66 geographic concepts that cover object- and scene-level categories. This arrangement follows the task setting in the main paper: the LR image supplies global context, while the HR tiles provide local evidence that can be acquired selectively.

\subsubsection*{A.2 Spatial and Temporal Alignment}
We initialize candidate scenes at the centroids of OSM polygons that satisfy the concept-filtering rules. A shared geographic footprint is defined around each anchor, retrieved from both sensors, and partitioned into the ordered HR grid. During index construction, we retain a candidate only if its full LR footprint falls within valid Sentinel-2 coverage and the corresponding HR grid intersects valid NAIP coverage. Both branches are then requested over the same geographic rectangle to preserve cross-scale spatial correspondence.

We also match the observations in time. For each region, we first retrieve the most recent NAIP image available between 2022-01-01 and 2024-04-01. We then search Sentinel-2 observations within a $\pm183$-day window around the NAIP acquisition date, constrained to 2023. If the window contains no valid observation, we fall back to the full 2023 range. We retain Sentinel-2 candidates with a reported cloudy-pixel ratio below 2\% and apply pixel-level masking from the scene-classification layer. These steps reduce geometric and temporal discrepancies before pretraining.

\subsubsection*{A.3 Quality Control}
We filter the data at both the metadata and image levels. During OSM parsing, we remove invalid geometry, insufficient spatial support, irrelevant tags, and near-duplicate anchors. We also discard candidates whose labeled area is too small relative to the scene footprint or whose NAIP coverage is invalid. After download, we screen LR and HR imagery for empty content using a black-pixel ratio threshold of 0.25. An invalid LR image removes the entire scene, while a missing, corrupted, or excessively empty HR tile removes the corresponding pair. A final consistency pass requires one valid LR image and all 100 ordered HR tiles in every retained scene.

\subsubsection*{A.4 Dataset Scale and Use}
After coverage, temporal, and image-quality filtering, GL-10M contains nearly 100,000 complete LR-to-HR scene pairs across 66 geographic concepts in 11 U.S. states. Each pair has one LR image and 100 local HR images, for approximately 10 million images in total. The training, validation, and test data are separated geographically, and the concept distribution is long-tailed. We use GL-10M for large-scale pretraining of the observation-guided latent predictor. As specified in the main paper, the Sentinel-2/NAIP models are fine-tuned, whereas Table~\ref{tab:zeroshot_recognition} directly tests the pretrained model on EuroSAT, BigEarthNet, and held-out GL-10M data without further fine-tuning. Table~\ref{tab:gl10m_summary} summarizes the retained data and release structure.

\begin{table*}[t]
  \centering
  \caption{Summary of the construction and released structure of GL-10M.}
  \label{tab:gl10m_summary}
  \small
  \setlength{\tabcolsep}{7pt}
  \begin{tabular}{ll}
    \toprule
    Property & Description \\
    \midrule
    LR imagery & Sentinel-2, 10\,m spatial resolution \\
    HR imagery & NAIP, 1\,m spatial resolution \\
    Pair structure & One LR overview and 100 ordered HR tiles on a $10\times10$ grid \\
    Semantic source & OSM-derived mapping to 66 geographic concepts \\
    Geographic coverage & 11 U.S. states with geographically separated data partitions \\
    Dataset scale & Nearly 100,000 scene pairs and approximately 10 million images \\
    Primary use & Large-scale pretraining of the observation-guided latent predictor \\
    \bottomrule
  \end{tabular}
\end{table*}

\subsubsection*{A.5 Construction Pipeline}
The construction pipeline maps OSM annotations to benchmark concepts, verifies Sentinel-2 and NAIP coverage, defines a shared LR footprint with a matching $10\times10$ HR grid, retrieves temporally matched imagery, and applies pair-level quality control. The resulting pairs preserve the cross-scale correspondence needed to learn from global LR context and selectively observed HR evidence.

\subsection*{B. Qualitative Analysis of LR-Guided Latent Completion}

\subsubsection*{B.1 Retrieval-based Diagnostic}
Figure~\ref{fig:appendix_completion} uses nearest-neighbor retrieval to inspect the completed representations. For each query scene, we apply the same sparse-observation mask to the HR tiles and inspect one held-out tile in the completed grid. Its pooled feature retrieves the three nearest neighbors from a fixed training-HR pool. All variants share the encoder, pooling rule, feature normalization, observed HR evidence, retrieval database, and similarity computation; only the completion route changes. The retrieved examples make semantic drift visible, while the main paper provides the quantitative evaluation.

\subsubsection*{B.2 Effect of LR Context}
\begin{figure*}[t]
  \centering
  \setlength{\tabcolsep}{1.5pt}
  \begin{tabular}{@{}c@{\hspace{0.002\linewidth}}c@{\hspace{0.002\linewidth}}c@{\hspace{0.002\linewidth}}c@{}}
    \includegraphics[height=0.40\textheight]{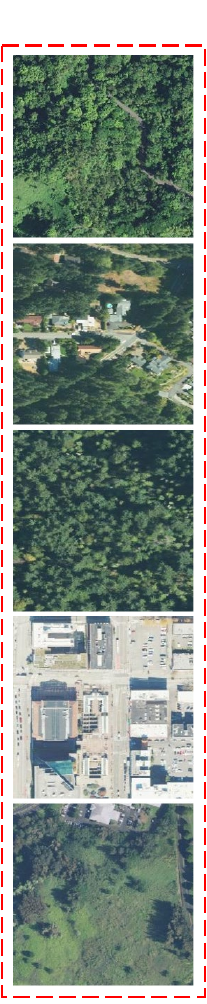} &
    \includegraphics[height=0.40\textheight]{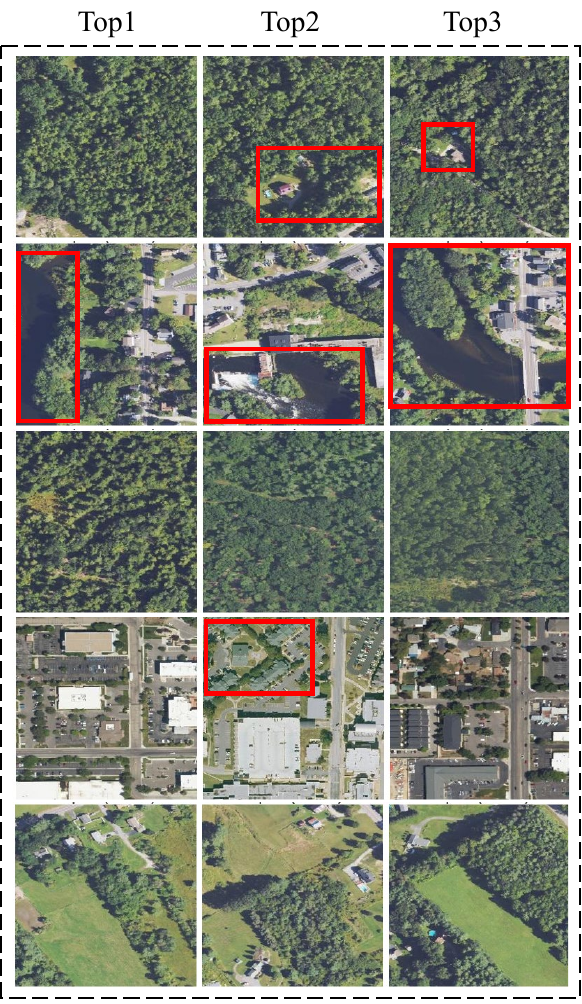} &
    \includegraphics[height=0.40\textheight]{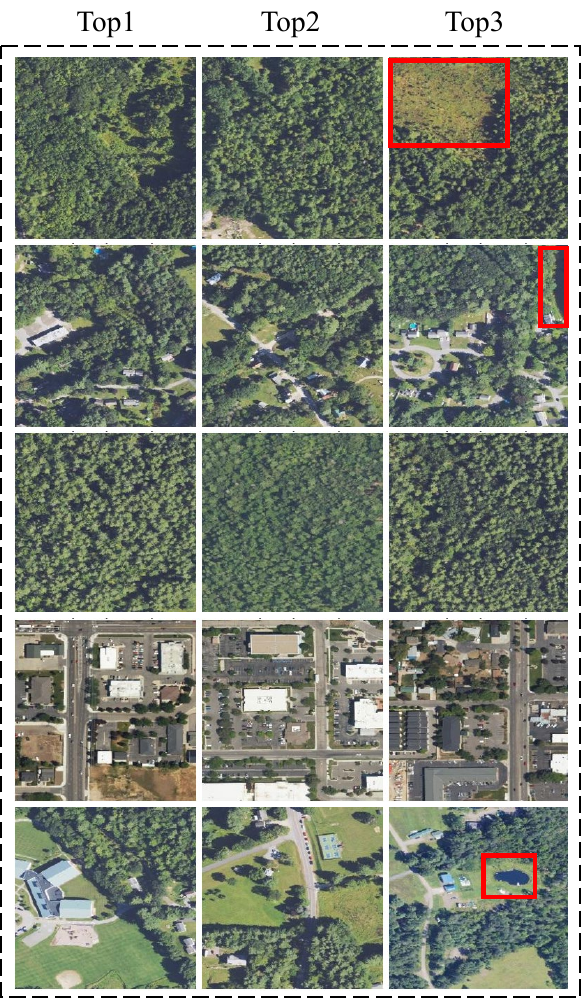} &
    \includegraphics[height=0.40\textheight]{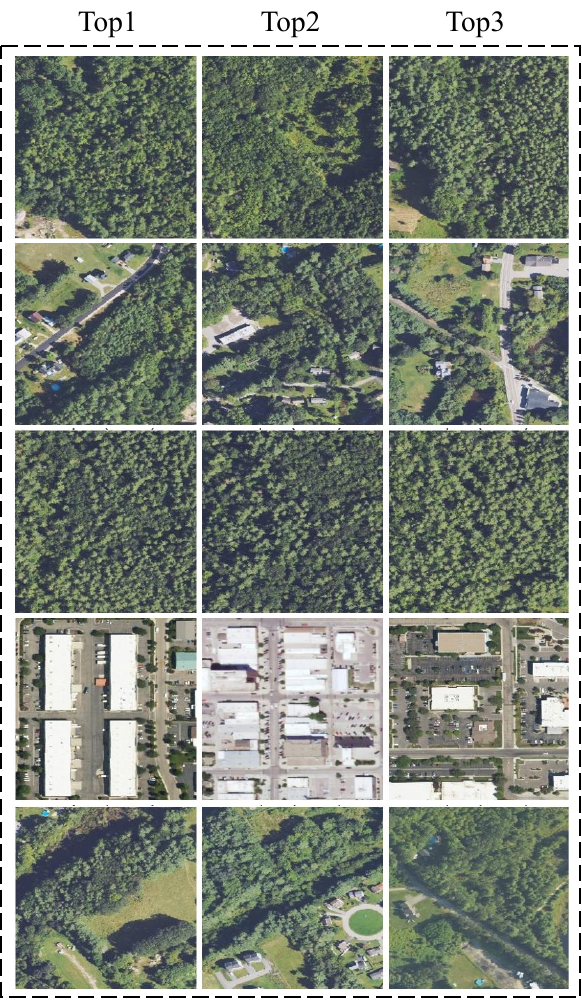} \\
    \small (a) Target tiles &
    \small (b) Without LR context &
    \small (c) With LR context &
    \small (d) GL-10M pretraining + LR context
  \end{tabular}
  \Description{Five target high-resolution tiles and their top-three nearest-neighbor retrievals under completion without low-resolution context, with low-resolution context, and with both GL-10M pretraining and low-resolution context. Red boxes mark visually inconsistent regions in selected retrievals.}
  \caption{\textbf{Retrieval-based diagnostic of latent completion.} The left column shows target HR tiles; the remaining groups show the top-three neighbors retrieved by features completed without LR context, with LR context, and with GL-10M pretraining plus LR context. Red boxes mark representative local mismatches.}
  \label{fig:appendix_completion}
\end{figure*}

Without LR context, the predictor must infer a missing tile from sparse local HR evidence alone. More than one scene interpretation may fit these observations, so retrieved neighbors can match local texture or color while differing in category or layout. The LR overview provides a scene-level reference for interpreting the sparse HR tiles. In Figure~\ref{fig:appendix_completion}, the LR-guided variants generally retrieve examples that better match the target structure. The GL-10M-pretrained variant also shows the effect of learning cross-scale regularities from a larger collection. LR context therefore constrains latent completion without replacing the selected HR evidence.

\subsection*{C. Qualitative Threshold-Sweep Analysis}

\subsubsection*{C.1 Sweep Setup}
\begin{figure*}[t]
  \centering
  \includegraphics[width=\textwidth,height=0.90\textheight,keepaspectratio]{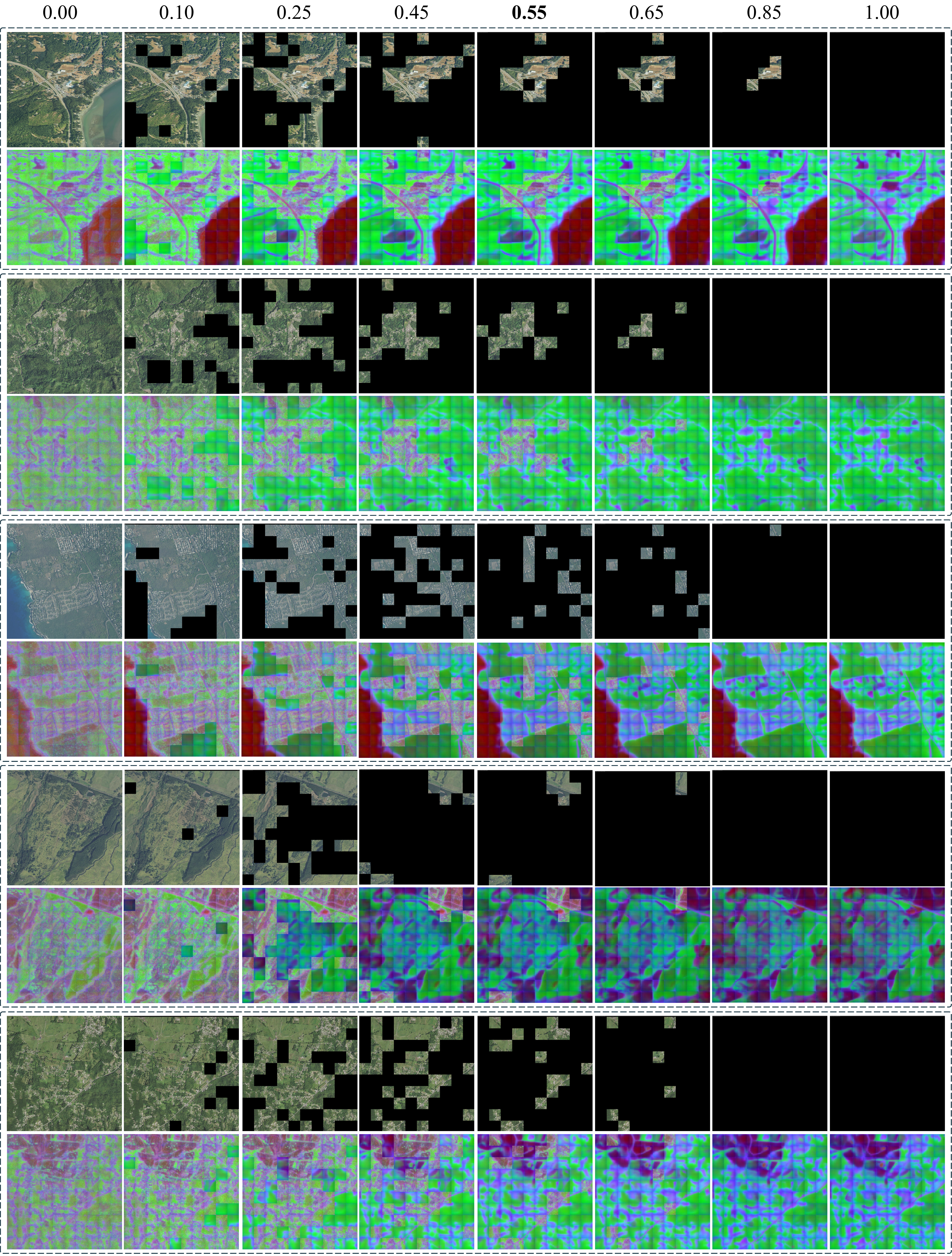}
  \Description{Five representative scenes under thresholds from 0.00 to 1.00. Each scene shows the retained high-resolution tiles and the corresponding completed latent visualization as the observation budget decreases.}
  \caption{\textbf{Qualitative threshold sweep under controlled HR acquisition.} The threshold increases from 0.00 (full HR observation) to 1.00 (no HR observation); 0.55 is the shared operating point used in the main experiments.}
  \label{fig:appendix_budget_completion}
\end{figure*}

Figure~\ref{fig:appendix_budget_completion} varies the shared sampling threshold across 0.00, 0.10, 0.25, 0.45, 0.55, 0.65, 0.85, and 1.00, with the rest of the inference pipeline fixed. Lower thresholds retain more HR tiles; higher thresholds reduce the realized observation budget. At 0.00, the complete HR grid provides the full-observation reference. At 1.00, no HR tiles remain, and the predictor relies on the LR overview and its learned completion prior. The main experiments use a shared threshold of 0.55.

\subsubsection*{C.2 Progression Across Thresholds}
Increasing the threshold progressively replaces direct HR evidence with predicted features. At low and intermediate thresholds, the main scene organization remains visually similar to the full-observation representation despite the smaller number of retained HR tiles. At high thresholds, local boundaries and small structures become less distinct, and LR context has a stronger influence on the completed features. The final column shows the limiting case with no HR observation. Across the examples, the transition is gradual, consistent with using the threshold to control the balance between acquired evidence and latent completion.

\subsubsection*{C.3 Interpretation}
The sweep shows the different roles of local and global evidence. Observed HR tiles anchor details at selected positions, while the LR overview provides a reference for filling the rest of the grid. Moderate thresholds preserve local anchors and keep completion constrained by direct observations. Very high thresholds remove most anchors, making small or ambiguous structures harder to recover. The visualization complements the matched-budget results by showing how the representation changes along the same control variable.

\subsection*{D. Additional Implementation Details}

\subsubsection*{D.1 Sampler Supervision Target}
We train the sampler without manual region annotations. Its offline target is a patch-level acquisition-priority map on the fixed $10\times10$ HR grid. Structural saliency estimates where fine spatial content is difficult to infer, while cross-scale semantic gain measures the evidence supplied by an HR tile beyond what is already present in the LR overview. Their Hadamard product gives high priority to locations supported by both cues. After global normalization, the resulting map $\mathbf{G}^{*}$ supervises the LR-to-score-map predictor with binary cross-entropy, as specified in the main paper. This target ranks candidate observations by their expected value under the acquisition budget, rather than assigning segmentation or class labels.

\subsubsection*{D.2 Separate Sampler Training}
We optimize the sampler and latent predictor in separate stages. Tile selection is discrete, and a strong predictor can partly compensate for weak acquisition choices. Downstream task supervision is therefore an indirect signal for the sampling policy. Offline supervision gives the sampler a direct target derived from paired LR and HR observations. Random budgeted masks expose the predictor to varied sparse-observation patterns before pairing it with the learned policy. This design gives each component a training signal suited to its role.

\subsubsection*{D.3 Threshold, Runtime, and Observation Cost}
At inference, a shared threshold of 0.55 is applied to globally normalized sampler scores. The same rule is used for every scene, but the number of selected HR tiles remains content-dependent. The reported Observation Budget Ratio (OBR) is therefore the realized dataset-level mean rather than a fixed per-scene quota. Under the equal-area $10\times10$ partition, OBR equals the requested HR area divided by the full HR area. Runtime measurements cover full-pipeline inference over the test set with batch size 32 on one NVIDIA RTX 4090 GPU, including sparse HR selection and downstream prediction.

\subsection*{E. Failure Cases and Limitations}
Figure~\ref{fig:failure_cases} groups three recurring failure patterns: missed informative HR regions, drift under very sparse evidence, and complex scene structure. They show where the assumptions behind selective observation and latent completion become restrictive.

\subsubsection*{E.1 Missed Informative HR Regions}
The sampler can under-allocate HR observations to small but decisive structures whose importance is weakly expressed in the LR overview. In Figure~\ref{fig:failure_cases}(a), only part of the relevant local structure is observed. The predictor must then complete the remaining area from incomplete evidence, and the missing detail persists in the resulting representation. This pattern is most likely when informative objects occupy a small fraction of the scene or differ only subtly from their surroundings at LR.

\subsubsection*{E.2 Drift Under Sparse Evidence}
Even with a plausible sampling pattern, completion can drift when the amount of observed HR evidence is small relative to scene complexity. Figure~\ref{fig:failure_cases}(b) shows a feature map that remains globally coherent but departs from locally supported structure. LR context narrows the set of plausible completions but cannot eliminate ambiguity when several local configurations agree with the same global overview.

\subsubsection*{E.3 Complex Scene Structures}
\begin{figure}[t]
  \centering
  \begin{minipage}[t]{0.33\columnwidth}
    \centering
    \includegraphics[width=\linewidth]{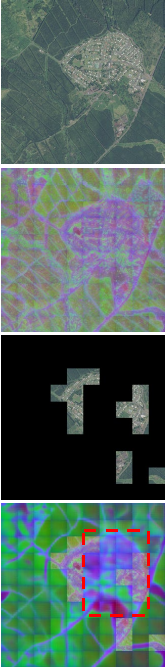}
    \small (a) Missed informative HR regions
  \end{minipage}\hfill
  \begin{minipage}[t]{0.33\columnwidth}
    \centering
    \includegraphics[width=\linewidth]{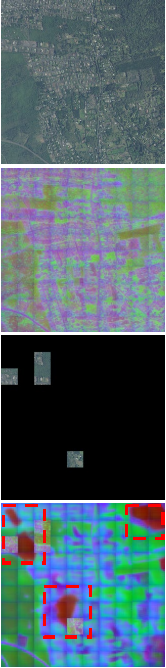}
    \small (b) Drift under sparse evidence
  \end{minipage}\hfill
  \begin{minipage}[t]{0.33\columnwidth}
    \centering
    \includegraphics[width=\linewidth]{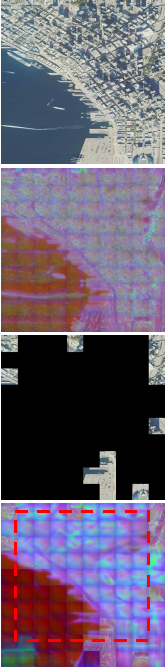}
    \small (c) Complex scene structures
  \end{minipage}
  \Description{Three examples containing an input scene, a latent feature visualization, sparse high-resolution evidence, and the resulting completion. Red dashed boxes mark representative regions associated with each failure pattern.}
  \caption{\textbf{Representative failure cases.} (a) missed informative HR regions, (b) drift under sparse evidence, and (c) complex scene structures.}
  \label{fig:failure_cases}
\end{figure}

Some scenes depend on long-range geometry, sharp boundaries, or rare local configurations. Figure~\ref{fig:failure_cases}(c) illustrates a case in which sparse observations do not preserve enough structural detail for reliable completion. Dense mixed-use regions, transportation hubs, and small objects embedded in clutter can therefore retain a larger gap from full-HR perception than more regular scenes.

\subsubsection*{E.4 Broader Limitations}
The framework assumes that a useful global LR observation is available and that HR observations can be requested on a predefined candidate grid. It also relies on the quality of the cross-scale training pairs and the stability of the sampler's offline supervision. Irregular observation geometry, larger temporal discrepancies, and stronger domain shift can violate these assumptions. Extensions could therefore consider adaptive partitioning, uncertainty-aware acquisition, and greater robustness to imperfect cross-scale correspondence.

\end{document}